\documentclass{article}

\usepackage{PRIMEarxiv}
\usepackage{amsmath} 
\usepackage{svg} 
\usepackage[utf8]{inputenc} 
\usepackage[T1]{fontenc}    
\usepackage{hyperref}       
\usepackage{url}            
\usepackage{booktabs}       
\usepackage{amsfonts}       
\usepackage{nicefrac}       
\usepackage{microtype}      
\usepackage{svg} 
\usepackage{svg} 
\usepackage{lipsum}
\usepackage{fancyhdr}       
\usepackage{graphicx}       
\usepackage{svg}            
\graphicspath{{media/}}     

\title{A Categorical Error Sensitivity Index (ISEC):\\
A Preventive Ordinal Decision-Support Measure for\\
Irrecoverable Errors in Manual Data Entry Systems}

\author{
  Mauro Anibal Benetti\\
  Universidad Tecnologica Nacional \\
  Facultad Regional San Rafael \\
  Mendoza, Argentina\\
  \texttt{mbenetti@frsr.utn.edu.ar} \\
   \And
  Ricardo Raúl Palma\\ 
  Universidad Nacional de Cuyo \\
  Instituto de Ingeniería \\
  Mendoza, Argentina\\
  \texttt{rpalma@uncu.edu.ar} \\
  \And
  Fabricio Orlando Sanchez Varretti\\
  Universidad Tecnologica Nacional \\
  Facultad Regional San Rafael \\
  Mendoza, Argentina\\
  \texttt{fsanchez@frsr.utn.edu.ar} \\
  }

\begin{document}
\maketitle

\begin{abstract}
Data-entry systems remain structurally vulnerable to categorical misclassifications, particularly in small and medium-sized enterprises (SMEs). When nominal categories exhibit semantic or morphological proximity, human–machine interaction may produce errors that are irrecoverable ex post. In the absence of automated input controls, manual data entry frequently generates irrecoverable categorical distortions that propagate into Key Performance Indicators (KPIs), thereby misleading managerial decision-making. 

State-of-the-art normalization tools typically evaluate semantic and morphological dimensions in isolation and rely heavily on standard dictionaries, rendering them ineffective for SME master data rich in custom SKUs, abbreviations, and domain-specific technical jargon. 

This paper introduces the Categorical Error Sensitivity Index (ISEC), an ordinal composite score designed to rank category pairs according to their structural susceptibility to confusion. ISEC integrates semantic distance (via word embeddings), custom-weighted morphological transformation costs (through an adapted Damerau–Levenshtein algorithm), and empirical frequency into a unified, mathematically robust preventive framework. By leveraging vector database architectures, ISEC reduces computational complexity, achieving approximately a 195× performance improvement over brute-force methods. Validated across three heterogeneous datasets: governmental judicial records, retail inventory, and a synthetic ISO-coded metalworking catalog, ISEC provides a scalable and proactive data governance instrument that enables SMEs to detect latent structural risk embedded within their categorical data assets.
\end{abstract}


\section{Introduction}

Small and Medium-Sized Enterprises (SMEs) operate under structural constraints in technological resources, data governance capacity, and formalized information management processes. In such environments, the human–machine interface remains the weakest link in the data ingestion process. Unlike large corporations with automated validation pipelines and structured master data governance, SMEs frequently rely on manual categorical inputs embedded in information systems, spreadsheets, and operational databases.

This structural reliance on manual data capture introduces a critical yet under-theorized vulnerability: categorical fragility \cite{unterkalmsteiner_2023}. Nominal categories, product codes, supplier names, incident types, labels are treated as stable semantic units. However, when entered manually, these categories become susceptible to typographical variations, morphological distortions, semantic ambiguity, and inconsistent abbreviations \cite{Feit2016,Navarro2001,vanderGoot2017}.

The consequences are not merely cosmetic. Certain categorical errors are \textbf{irrecoverable}. Once an incorrect category is stored and propagated across aggregation layers, it becomes statistically indistinguishable from legitimate entries \cite{Cichy2019}. When traditional data cleaning techniques cannot reconstruct the intended meaning, the error may becomes structurally embedded in the informational architecture.

This paper argues that categorical sensitivity to error is not a downstream data quality issue \cite{Wand1996}, but a measurable property of system design. We introduce the \textbf{Index of Sensitivity to Categorical Errors (ISEC)}, an ordinal metric designed to rank nominal taxonomies according to their latent vulnerability to confusion prior to error occurrence \cite{lorena_2019}.

ISEC shifts data quality management from reactive correction toward preventive risk modeling \cite{Cichy2019,Miller2025}, particularly at the level of Decision Support Systems (DSS), where aggregated categorical distortions may produce structured misinformation \cite{Ghasemaghaei2018,MendesSampaio2015,Shrestha2021,Velden2024}.

\section{Problem Statement and Situational Context}

\subsection{Structured Misinformation and Algorithmic Indeterminacy}

Decision Support Systems (DSS) rely heavily on aggregated nominal data to compute Key Performance Indicators (KPIs) \cite{MendesSampaio2015}. If the underlying taxonomy is fragile, aggregation may produce \textit{structured misinformation}: internally consistent metrics derived from semantically unstable inputs \cite{Simard2019,Wang1996}. 

Categorical fragility manifests critically in an operational regime where the effective structural separation between two valid categories is naturally short \cite{lorena_2019,unterkalmsteiner_2023}. Crucially, an error becomes \textbf{irrecoverable} not necessarily because the structural distance between the intended label and another label drops to zero. Rather, an error is irrecoverable when a small stochastic perturbation (a typographical error) places the erroneous entry equidistant between two valid categories, or strictly closer to an incorrect one \cite{Navarro2001}. 

If the edit distance from the erroneous item to two different valid 
categories is the same, and there is no significant difference in their 
empirical frequencies, the classification decision ceases to be strictly deterministic and becomes probabilistically indeterminate \cite{Aliero2023}.

\begin{figure}
    \centering
    \includegraphics[width=1\linewidth]{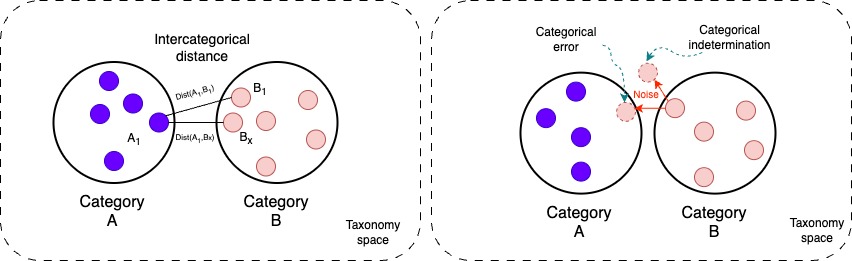}
    \caption{Inter-categorical distance between categories and the effect of errors on the capture process.}
    \label{fig:placeholder}
\end{figure}

Once this state of structural ambiguity is reached, systems operating 
solely on internal similarity measures lack sufficient discriminatory 
power to deterministically assign the correct category \cite{Malkov2020,Navarro2001}.

\subsection{Limitations of Current Metrics: The Compartmentalization of Dimensions}
The literature reveals a critical gap in the treatment of non-standard words (e.g., SKUs, industry acronyms, and alphanumeric ISO codes) \cite{Aliero2023,Miller2025}. State-of-the-Art (SOTA) text normalization systems, demand massive computational power and extensive training corpora \cite{Aliero2023,Johnson2024,Shrestha2021} resources unavailable to most SMEs and specially the Argentinean SME \cite{Alsousi2022,Gnther2019}. 

Furthermore, traditional metrics suffer from the "compartmentalization of dimensions" \cite{Cichy2019,Wang1996}. They evaluate morphological distance (how a word is written) and semantic distance (what a word means) independently \cite{Johnson2024,MendesSampaio2015,Po2020,unterkalmsteiner_2023}. Standard spell-checkers evaluate structural edits but ignore the semantic plausibility of the correction \cite{Po2020,Zhao2019}, while semantic models fail to account for the physical ergonomics of typing errors \cite{Feit2016}. ISEC bridges this gap by unifying these dimensions into a single actionable index.

\subsection{Irrecoverable Errors}

An error is considered irrecoverable when:

\begin{enumerate}
\item It produces a syntactically valid category.
\item It does not violate database constraints.
\item It remains undetected during aggregation \cite{Simard2019}.
\item The original intended label cannot be reconstructed ex post \cite{Aliero2023,Berger2021,vanderGoot2017}.
\end{enumerate}

Such errors differ fundamentally from missing data or format violations \cite{Aliero2023,Cichy2019,Feit2016,Miller2025}. They are epistemic distortions embedded within the classification layer.

\subsection{Human–Machine Interface Weakness}

SMEs depend heavily on manual UI forms without adaptive validation or semantic distance monitoring \cite{GohZhiJi2023,Gnther2019,Shrestha2021}. Standard spell-checkers are ineffective because SMEs master data often include \cite{Aliero2023,Miller2025,vanderGoot2017}:

\begin{itemize}
\item SKUs
\item Technical nomenclatures
\item Industry-specific acronyms
\item Non-standard abbreviations
\item Alpha-numeric codes
\item Evolving names and abbreviations
\end{itemize}

Thus, the interface itself becomes a structural generator of latent informational entropy \cite{lorena_2019,Wand1996}.
\section{Theoretical Positioning}

ISEC integrates four conceptual domains; Information Quality Theory, Linguistic Distance Modeling, Error Propagation in DSS, Organizational Risk Modeling. Rather than defining quality as conformance to specification ex post \cite{Wand1996,Wang1996}, we conceptualize quality as \textbf{ex ante structural robustness}. A taxonomy with high categorical proximity is intrinsically fragile regardless of whether errors have already been observed \cite{lorena_2019,unterkalmsteiner_2023}.

\subsection{Distance Compression and Probabilistic Collapse}

We define \textbf{Distance Compression under Stochastic Noise} to the phenomenon where small structural separation between categorical labels becomes further reduced through probabilistic typographical perturbations, leading to irrecoverable errors in the capture process of data systems.

It is important to clarify that categorical sensitivity does not arise merely because two categories exhibit short structural distance. Proximity alone does not constitute an error. Rather, sensitivity emerges when an already small inter-category distance is exposed to stochastic perturbations induced by manual data entry \cite{Feit2016,Wand1996,Zhao2019}.

Let $d(c_i,c_j)$ denote the composite structural distance between two categories. Consider a random perturbation operator $\epsilon$ representing typographical noise, abbreviation variability, or human transcription error \cite{Aliero2023,vanderGoot2017,Feit2016}. The effective post-perturbation distance becomes:

\[
d'(c_i,c_j) = d(c_i,c_j) - \epsilon
\]

where $\epsilon$ is a non-negative stochastic variable bounded by the maximum transformation cost induced by a capture variability \cite{Feit2016,lorena_2019,Navarro2001}.

When $d(c_i,c_j)$ is sufficiently small, exists a non-zero probability that the effective discriminative margin need it by the algorithm between $c_i$ and $c_j$ falls below the admissible threshold $\delta$ \cite{lorena_2019}:

\[
d'(c_i,c_j) \leq \delta
\]

From the perspective of a normalization or correction algorithm, both categories may become comparably plausible candidates within the tolerance bounds of the correction mechanism \cite{Aliero2023,Zhao2019}. The classification decision ceases to be strictly deterministic and becomes probabilistically indeterminate \cite{Simard2019}.

Thus, categorical fragility \cite{lorena_2019,unterkalmsteiner_2023} is not simply a function of proximity, but of \textbf{proximity under perturbation} when some perturbations are more likely than others \cite{Feit2016,lorena_2019}. The smaller the initial structural separation, the lower the perturbation threshold required to generate classification ambiguity. This phenomenon is what we termed \textit{distance compression under stochastic noise}. 


\begin{figure}
    \centering
    \includegraphics[width=1\linewidth]{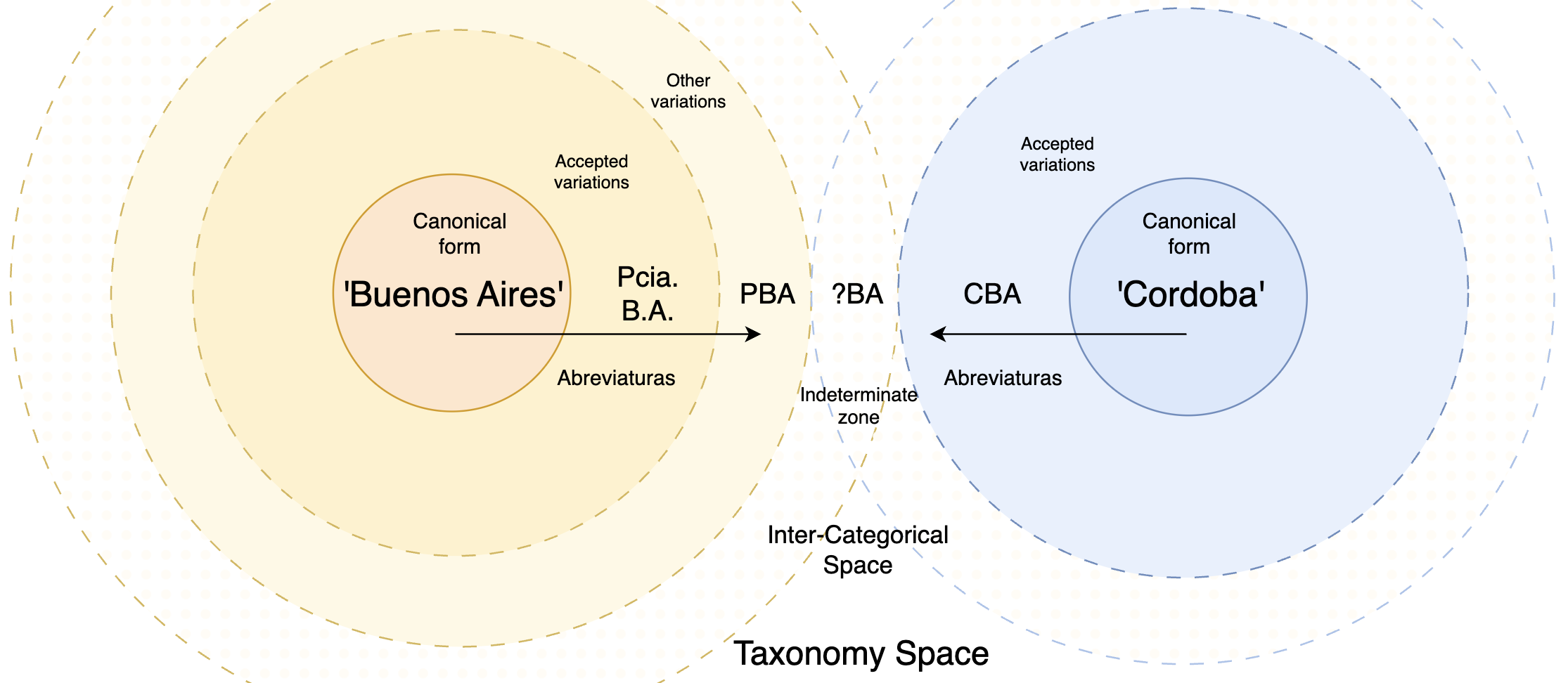}
    \caption{Example of a indeterminate zone between two categories when the first character of an abbreviation form is affected. Extracted from case 3, the taxonomy contains 30 canonical names based on Argentine provinces.}
    \label{fig:placeholder}
\end{figure}

\section{Formal Definitions}
\subsection{ISEC Definition}

The Inter-Categorical Sensitivity Index for a pair of categories 
$(c_i,c_j)$ is defined as:

\begin{equation}
ISEC_{(i,j)} =
\frac{1 + FMN_{(i,j)}}
{DSN_{(i,j)}^{\alpha} \times CMP_{(i,j)}^{1-\alpha}}
\end{equation}

Where:

\begin{itemize}
\item $FMN_{(i,j)}$ is the normalized mean frequency of the pair
\item $DSN_{(i,j)}$ is the normalized semantic distance
\item $CMP_{(i,j)}$ is the penalized mean transformation cost, computed 
using a weighted Damerau--Levenshtein distance in which each edit 
operation is assigned a cost according to a custom pair-wise 
character transformation matrix.
\item $\alpha \in [0,1]$ is a tuning parameter that determines the semantic vs. morphological weight. $\alpha \to 1$ prioritizes meaning (better for long descriptions), while $\alpha \to 0$ prioritizes structure (better for short SKUs).
\item $ISEC_{(i,j)} \in \mathbb{R}^+$
\end{itemize}

ISEC is an unbounded positive real-valued index used exclusively for ordinal ranking of categorical sensitivity. It does not represent a probability measure.

\vspace{2mm}

\subsection{Domain of ISEC}

Let $\mathcal{C} = \{c_1, c_2, ..., c_n\}$ be a finite set of nominal 
categories with $n \geq 2$.

The Inter-Categorical Sensitivity Index is defined over the domain:

\[
\mathcal{D} =
\{ (c_i,c_j) \in \mathcal{C} \times \mathcal{C} \mid i \neq j \}
\]

that is, over ordered pairs of \textbf{distinct categories}.

For every $(c_i,c_j) \in \mathcal{D}$, assume:

\[
DSN_{(i,j)} > 0
\quad \text{and} \quad
CMP_{(i,j)} > 0
\]

These strict positivity conditions exclude trivial identity cases 
and prevent denominator degeneracy. In particular:

\begin{itemize}
\item $DSN_{(i,j)} = 0$ would imply perfect semantic equivalence,
\item $CMP_{(i,j)} = 0$ would imply zero-cost structural transformation.
\end{itemize}

Both cases contradict the assumption $i \neq j$ and are therefore 
excluded from the admissible domain.

\subsection{FMN (Normalized mean frequency of the pair)}

This corresponds to the average frequency of occurrence of the two categories analyzed:

\[
\mathrm{FMN}_{(i,j)} = \log_{10}\!\left(\mathrm{FM}_{(i,j)}\right)
\]

Where:

\[
\mathrm{FM}_{(i,j)} = \frac{F_i + F_j}{2}
\]

The base-10 logarithm mitigates highly skewed distributions.
Typical values of the $\mathrm{FM}$ (average frequency of the pair)
range between $1$ and $10^{5}$ in datasets of medium size.

\vspace{5mm}

\subsection{DSN (Normalized Semantic Distance)}

It represents the semantic dissimilarity between categories, obtained as the
complement of the cosine similarity between their \textit{vector embeddings} \cite{Xie2023}.
Its range is between $0$ (semantically identical categories) and $1$ (highly dissimilar categories).

\[
\mathrm{DSN}_{(a,b)} = \frac{1 - \mathrm{CosineSimilarity}_{(a,b)}}{2}
\]

Where:

\[
\mathrm{CosineSimilarity}_{(a,b)} =
\frac{\mathbf{a} \cdot \mathbf{b}}
{\|\mathbf{a}\| \, \|\mathbf{b}\|}
\in [-1,1]
\]

\subsection{Penalized Mean Cost (CMP)}

The Penalized Mean Cost for a pair $(c_i,c_j)$ is defined as:

\[
CMP(c_i,c_j) = CM(c_i,c_j) + k \times CP(c_i,c_j)
\]

where $k \geq 0$ regulates the relative influence of cumulative 
linear structural divergence.

\vspace{3mm}
\textbf{Mean Transformation Cost (CM).}

Let the transformation of $c_i$ into $c_j$ require a sequence of $N$ weighted edit operations under a customized Damerau--Levenshtein scheme. Let $c_k$ denote the cost of the $k$-th operation as specified by the shared pair-wise character cost matrix. The mean transformation cost is defined as:

\[
CM(c_i,c_j) = \frac{1}{N} \sum_{k=1}^{N} c_k
\]

Thus, $CM$ represents the average weighted cost of all edit operations, including insertions, deletions, substitutions, and transpositions \cite{Zhao2019}.

\vspace{2mm}
\textbf{Penalty Cost (CP).}

We design CP to accumulate only insertions, deletions, and substitutions. Transpositions are excluded to preserve linear growth in string-length divergence, a modification of the classical Damerau-Levenshtein framework , which treats all four operations symmetrically \cite{Zhao2019}.
The linear penalty component isolates cumulative divergence associated exclusively with classical Levenshtein operations \cite{Po2020,Navarro2001}:

\[
CP(c_i,c_j) =
\sum_{L \in \mathcal{L}} w_L \cdot n_L(c_i,c_j)
\]

Where: \[\mathcal{L} = \{\text{insertion}, \text{deletion}, \text{substitution}\}
\]

Here, $w_L$ and $n_L(c_i,c_j)$ are derived from the same custom cost matrix used in $CM$. We introduce a linear penalty component $CP$ that excludes transpositions. Unlike classical Damerau-Levenshtein formulations \cite{Zhao2019} which treat transpositions as a single operation of unit cost, our $CP$ accumulates only insertions, deletions, and substitutions. This design choice preserves strictly linear growth with respect to string-length divergence, whereas including transpositions would introduce a non-linear compression artifact when transpositions co-occur with deletions or insertions.

Conceptually, $CM$ captures the average morphological effort required to transform one category into another, while $CP$ captures the total accumulated linear structural deviation. The penalization term 
$k \times CP$ amplifies cases in which divergence is driven by numerous operations.

\subsection{Behavioral Properties of CMP}

The Penalized Mean Cost $CMP(c_i,c_j)$ exhibits distinct growth regimes depending on the number and nature of transformation operations \cite{Berger2021}.

Let $N$ denote the total number of edit operations required to transform $c_i$ into $c_j$, and let $w_k$ be the weight associated with operation type $k$.

Two structural regimes may be identified:

\paragraph{1. Low-Operation Regime:} If $N$ is small and the associated weights $w_k$ are low, then:

\[
CMP(c_i,c_j) \downarrow
\]

In this regime, the effective morphological separation between categories is reduced but does not vanish. The structural distance remains strictly positive; however, its magnitude becomes sufficiently small that modest 
stochastic perturbations may significantly compress the residual discriminative capacity.

Thus, the system approaches a state of elevated categorical fragility without implying exact structural equivalence. The risk lies not in distance annihilation, but in distance compression toward a threshold 
at which probabilistic ambiguity becomes non-negligible.

\vspace{4mm}

\paragraph{2. High-Operation or High-Cost Regime:}

If either $N$ grows large, or the associated weights $w_k$ are default (e.g., deletion-dominant transformations), then:
\[
CMP(c_i,c_j) \uparrow
\]

Under these conditions, structural separation increases due to the penalization component:
\[
CMP(c_i,c_j) = CM(c_i,c_j) + k \times CP(c_i,c_j)
\]

Since $CP(c_i,c_j)$ accumulates linearly over insertions, deletions, and substitutions, the penalized term amplifies transformation magnitude as operation counts increase.

\vspace{2mm}

\begin{figure}[hbt!]
    \centering
    \includegraphics[width=1\linewidth]{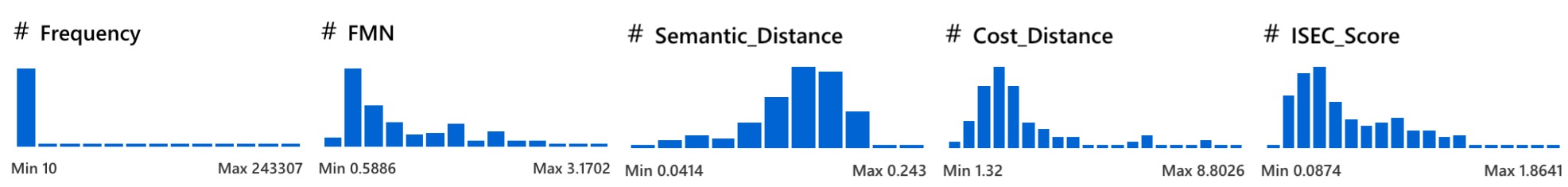}
    \caption{Illustrative distributions of the input profiles across the dimensions considered in the ISEC calculation, together with the resulting ISEC output profile for Case 3.}
    \label{fig:placeholder}
\end{figure}

\paragraph{Structural Interpretation:}Few low-cost operations reduce effective morphological cost and 
increase vulnerability to probabilistic collapse. Conversely, many operations or high-cost substitutions indicate structural separation, thereby increasing resilience against stochastic perturbation \cite{Berger2021,Po2020,Navarro2001}.

Thus, CMP behaves as a distance amplification mechanism under operation accumulation and as a compression indicator under low-cost minimal transformation paths. The CMP was designed to behave like the phenomenon introduced earlier, a distance that compress under stochastic bounded noise but expand when the number of transformations increase.

\subsection{Transformation Cost Matrices and Operational Weighting}

The transformation cost between two categorical labels depends not only on the number and type of edit operations required, but also on the specific characters involved in each operation \cite{Po2020,Navarro2001,Zhao2019}.

Rather than assigning a single scalar weight per operation type, ISEC implements a set of operation-specific cost matrices. For example:
\begin{itemize}
\item $\mathbf{W}^{(\text{sub})}_{a,b}$ represents the substitution cost 
of replacing character $a$ with character $b$.
\item $\mathbf{W}^{(\text{ins})}_{a}$ represents the cost of inserting 
character $a$.
\item $\mathbf{W}^{(\text{del})}_{a}$ represents the cost of deleting 
character $a$.
\item $\mathbf{W}^{(\text{trans})}_{a,b}$ represents the cost of 
transposing adjacent characters $a$ and $b$.
\end{itemize}

Each matrix contains:

\begin{itemize}
\item A default baseline cost for unspecified character pairs ($w_k = 1$) falling under a classic edition distance operation.
\item Optional custom overrides defined via a configuration file.
\end{itemize}

This architecture allows domain-specific calibration. For instance:

\begin{itemize}
\item Adjacent keys on a QWERTY keyboard (e.g., G $\leftrightarrow$ T) may receive reduced substitution cost \cite{Feit2016}.
\item Frequently confused numeric characters (e.g., 0 and O, 1 and I) may be assigned lower costs.
\item Transpositions may be assigned lower cost by default \cite{Berger2021,Zhao2019}.
\end{itemize}

Under the default symmetric configuration, all matrices collapse to unitary costs, reducing the metric to the classical Damerau-Levenshtein formulation \cite{Zhao2019}.

However, the matrix-based design enables fine-grained modeling of morphological risk, allowing organizations to encode empirical error tendencies and device-specific input behaviors directly into the distance computation \cite{Feit2016}.

\subsubsection*{Operational Interpretation of the Cost Matrices and $k$}

The transformation cost matrices and the penalization parameter $k$ are not merely computational hyper-parameters; they can encode organizational risk preferences as well as the particular capture process within the information system \cite{Wang1996, Cichy2019}.

\paragraph{Penalization Parameter $k$:}

While the cost matrices model micro-level perturbation behavior, the parameter $k$ governs macro-level amplification of morphological divergence through the linear penalty component:

\[
CMP(c_i,c_j) = CM(c_i,c_j) + k \times \, CP(c_i,c_j)
\]

Operationally:

\begin{itemize}
\item If $k \approx 0$, the system emphasizes average per-operation severity, allowing moderate structural flexibility.
\item If $k$ increases, cumulative insertions, deletions, and substitutions are increasingly amplified, expanding effective separation between categories.
\end{itemize}

Thus, $k$ controls how aggressively the system penalizes structural accumulation. It acts as a curvature regulator over distance growth, determining whether the system tolerates gradual lexical drift or rapidly escalates separation under repeated edits.

\paragraph{Governance-Level Calibration:}

The character-level cost matrix is initialized with a default unitary cost ($w_k = 1$) for all edit operations, corresponding to the classical symmetric implementation of the Damerau--Levenshtein distance \cite{Berger2021,Zhao2019}. 

The configuration file allows selective cost reductions for transformations considered more probable or operationally low-impact. For example, substitutions between characters that are adjacent in the QWERTY keyboard (e.g., ``T'' and ``G'') may be assigned a lower weight to reflect their higher likelihood during manual typing \cite{Feit2016}.

Because $CMP$ appears in the denominator of the ISEC formulation, reducing the cost of a transformation decreases $CMP$ and therefore increases the resulting sensitivity score. In this way, lower weights do not indicate reduced risk; rather, they model higher structural susceptibility to confusion under realistic 
human–machine interaction conditions.

Consequently, calibration of the cost matrix and $k$ defines an explicit governance surface that formalizes organizational assumptions about typographical probability and categorical robustness \cite{Wang1996,Cichy2019,lorena_2019,unterkalmsteiner_2023}.

To support empirical calibration and reproducibility, the public ISEC repository \footnote{https://github.com/mbenetti/isec} includes an interactive web application that enables experimentation with alternative cost matrices and $k$ configurations . The repository also provides a default configuration file corresponding to the classical unit-cost baseline, along with adjusted parameter sets that performed robustly in the empirical use cases presented in this study.

This interactive environment allows practitioners to explore how different probabilistic typing assumptions alter categorical sensitivity rankings, thereby making parameter selection a transparent and evidence-based governance decision.

\section{Implications for Data Quality and Information Systems}

ISEC operates at the pre-aggregation layer \cite{Cichy2019,Simard2019}. Its strategic relevance lies in:

\begin{itemize}
\item Identifying fragile taxonomies before or after deployment \cite{lorena_2019, unterkalmsteiner_2023}
\item Supporting UI/UX redesign decisions \cite{unterkalmsteiner_2023,Wang1996}
\item Guiding category consolidation if necessary \cite{lorena_2019, unterkalmsteiner_2023}
\item Preventing propagation of irrecoverable distortions
\item Provide additional metadata about the underlying information used for any KPI construction in particular \cite{Ghasemaghaei2018,MendesSampaio2015,Shrestha2021}
\end{itemize}

Unlike post-hoc audits, ISEC models \textbf{latent structural risk} \cite{lorena_2019,Miller2025,Simard2019}. This is particularly relevant for SMEs, where resource constraints prevent continuous data cleansing cycles.

\section{Computational Optimization done by ISEC}

A brute-force calculation of distances for $N$ categories results in $O(N^2)$ complexity \cite{Zhao2019,Navarro2001}, making it nonviable for catalogs exceeding a few thousand items \cite{Berger2021}. 

To ensure technological sovereignty and low implementation costs \cite{Gnther2019, Navarro2001,vanderGoot2017}, ISEC is architected for open-source stacks using a \textbf{Hybrid Search Strategy} via Vector Databases (e.g., ChromaDB with HNSW algorithms \cite{Malkov2020,Xie2023}) using two strategies:

\begin{enumerate}
    \item \textbf{Breadth-First (Semantic):} The implementation rely on VectorDB queries that according to estimations to retrieve the Top-$K$ semantic neighbors for a given item the cost is approximately $O(\log N)$ time \cite{Malkov2020,Xie2023}.
    \item \textbf{Greedy (Morphological):} The $CMP$ is strictly calculated against these $K$ candidates in $O(K)$ time. When Top-K is equal to the number of categories all categories will be measure reducing the hybrid search to a simple greedy search
\end{enumerate}
This reduces overall complexity to $\mathbf{O(\log N + k)}$. In empirical testing with 1,000 categories, processing time dropped from 8.4 hours to \textbf{2.6 minutes} an efficiency achieving approximately a 195× performance improvement over brute-force methods \footnote{https://github.com/mbenetti/isec.git}.

\section{Empirical Validation Across Heterogeneous Domains}

To evaluate the practical applicability of ISEC under real-world constraints, the metric was validated across three heterogeneous datasets, originally presented in a doctoral research . The objective was not merely to detect existing errors, but to assess the capacity of ISEC to identify structural categorical sensitivity prior to normalization \cite{Wang1996,Cichy2019}.

All experiments were executed on a virtual machine equipped with 4 vCPUs ($2$ CPU cores at $3.8$ GHz) and $8$ GB of RAM, without GPU acceleration. The ISEC implementation used approximately 2.7 GB of RAM, of which 1 GB was allocated to the embedding model and the remainder to the Python execution environment

\subsection{Case 1: Synthetic ISO-Based Dataset}

The first validation scenario employed a controlled synthetic dataset structured according to ISO 1832 \footnote{ISO 1832:2017 https://standards.iteh.ai/catalog/standards/sist/af10a357-79be-4c54-b963-59800aeb8545/iso-1832-2017}, the international standard defining nomenclature for indexable inserts in metal cutting tools. While the standard allows for nearly 100 million theoretically unique combinations, SMEs typically operate with a much smaller subset. A list of 1,000 valid codes was generated to test the index. Because this simulated a new system without historical usage, the empirical frequency ($FMN$) for all items was set to 1, isolating the purely structural risk

This environment allowed manipulation of:
\begin{itemize}
\item Category spacing.
\item Frequency distribution of the characters and numbers used for generation.
\end{itemize}

ISEC processed the 1,000 categories in 2 minutes utilizing a balance parameter of $\alpha=0.4$ and a Top-K=10 semantic search \cite{Malkov2020}. The morphological calculation utilized a custom substitution cost matrix reflecting physical typing habits on a QWERTY keyboard \cite{Feit2016}; adjacent keys (e.g., G $\leftrightarrow$ T, K $\leftrightarrow$ L) were assigned a reduced penalty cost in comparison to distant keys. The selection of which keys are in proximity was arbitrary and shows the flexibility of the index to be adapted to the capture process.

The controlled setup confirmed the theoretical proposition of probabilistic distance collapse \cite{Berger2021,Po2020,Navarro2001}. ISEC identified "silent" vulnerabilities where minor typographical noise generated entirely different but completely valid codes. For example, the code `AAGX110216` was highly sensitive to transforming into `AGAX110216` due to low-cost transpositions \cite{Berger2021,Zhao2019}.

\subsection{Case 2: Retail Supermarket Inventory}

The second case examined was a retail SME inventory database comprising 25,638 manually entered product descriptions distributed across 14 ontological groups (e.g., 'Aseo de hogar', 'Cuidado Personal', 'Charcutería'). Unlike the first case, this environment exhibited long descriptive strings (sentences) with high lexical variability and semantic density \cite{Aliero2023}. The dataset can be found in \footnote{https://www.kaggle.com/datasets/camiloemartinez/productos-consumo-masivo}

The analysis focused on an inter-group comparison, evaluating items from one group against items in all other groups to detect latent overlaps yielding a massive search space of 297,404,181 unique pairs. The ISEC implementation allows to perform "same group" or "different group" comparisons. Applying traditional brute-force computational methods to calculate semantic and morphological distances for this volume, considering our setup, would require approximately 187 days of processing time. However, leveraging ISEC's optimized hybrid architecture (combining Vector Database ANN search \cite{Malkov2020,Xie2023} with bounded morphological processing), the processing time was reduced to just \textbf{12 minutes} \footnote{https://github.com/mbenetti/isec.git}.

The ISEC successfully revealed structural duplicates and semantic collisions spanning different classification groups. Some of the critical vulnerabilities identified were:

\begin{itemize}
\item ``Bizcocho achiras del Huila'' assigned to the \textit{Panadería y Pastelería} group vs. ``Bizcocho achiras del huila'' assigned to \textit{Dulces y postres}.
\item ``Chicharrón Espirales'' in \textit{Charcutería} vs. ``Chicharrón espirales'' in \textit{Pasabocas}.
\item ``Quitamanchas Fab ropa blanca líquido'' in \textit{Aseo de hogar} vs. ``Quitamanchas Fab ropa color blanca líquido'' in \textit{Cuidado de ropa y calzado}.
\end{itemize}

These and other findings empirically confirm the claim that semantic embeddings alone are insufficient for quality assessment when naming conventions share critical structural tokens across domain boundaries \cite{Johnson2024,Reimers2019}. In addition, customized penalized morphological costs \cite{Berger2021,Zhao2019} allowed the ISEC to surface the most affected categories by lower cost transformations.

\subsection{Case 3: Judicial Administrative Records (CAJ – Argentina)}
The last validation scenario utilized open-access data from the \textit{Centro de Acceso a la Justicia (CAJ)}, an Argentine Governmental legal assistance network. The dataset consisted of $1,069,280$ records across $32$ variables, collected between 2016 and 2019\footnote{https://datos.gob.ar/it/dataset/justicia-consultas-efectuadas-centros-acceso-justicia--caj-}.

The ISEC identified extreme vulnerabilities in short, high-frequency abbreviations. For instance, \textit{"caba"}, the abbreviation for \textit{"Ciudad Autónoma de Buenos Aires"} and \textit{"cba"} (Córdoba) differ by a single character. The pair \textit{"cba"} also heavily conflicted with other abbreviations like \textit{"pba"}, \textit{"gba"}, and \textit{"ba"}. The combined frequency of these highly sensitive abbreviations exceeded $56,000$ records (accounting for $6-7$\% of total entries). It was flagged that a single typo in these abbreviations causes irrecoverable  misclassification or loss of the datapoint given the irreversibility of the change. Identifying this vulnerability allowed for proactive recommendations such as targeted autocomplete suggestions or requesting a confirmation from the user about the intention for the datapoint when any of those abbreviations are used.

\begin{figure}
    \centering
    \includegraphics[width=1\linewidth]{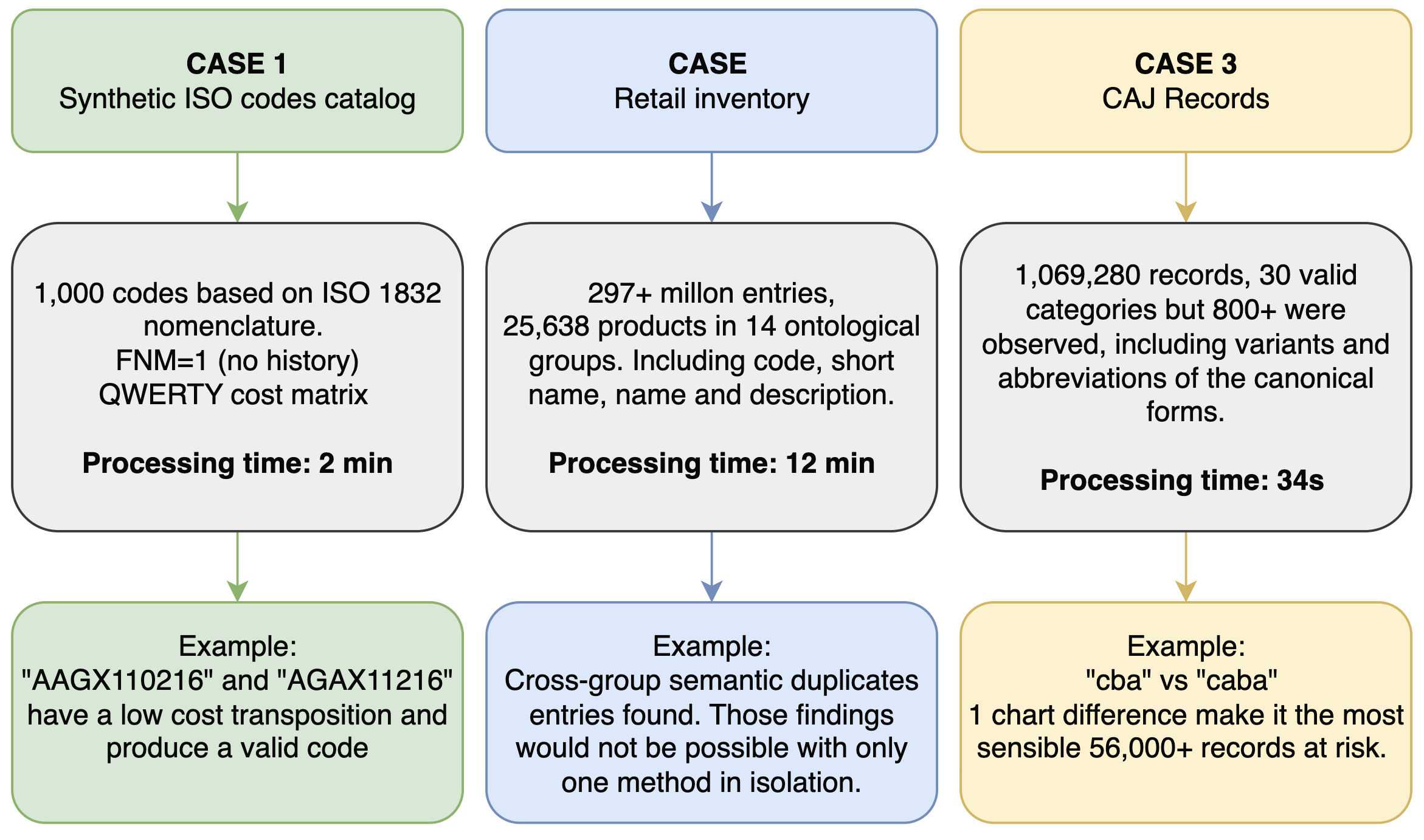}
    \caption{Summary of the index empirical application to three heterogeneous datasets}
    \label{fig:placeholder}
\end{figure}

\subsection{Cross-Case Observations}

Across the three domains, several structural regularities emerged:

\begin{enumerate}
\item High-frequency categories amplify distortion impact \cite{Cichy2019}.
\item Abbreviation systems significantly reduce effective inter-category distance \cite{Po2020,Navarro2001}.
\item Semantic embeddings or morphologic distances alone are insufficient for discriminability \cite{Johnson2024,Reimers2019,Velden2024}.
\item When empirical frequency data is unavailable, as in pre-deployment catalogs, the index operates exclusively on structural proximity, which is precisely the intended behavior: categories that are morphologically or semantically close are structurally vulnerable regardless of how often they appear, since a single typographical error may render them mutually irrecoverable. Frequency weighting, when available, further prioritizes pairs whose operational exposure amplifies the impact of that structural vulnerability.
\item The ISEC implementation allows to adapt the cost matrix to any distribution and to balance the semantic or morphological impact accordingly to the scenario; numeric or alphanumeric codes, short medium and long category names, long sentences or even paragraphs are also possible to be process and determine the sensitivity between those categories.
\end{enumerate}

Most critically, in all three cases, ISEC identified high-risk categorical zones before normalization procedures were applied, confirming its preventive capacity \cite{Wang1996,Cichy2019}.

These results support the central claim of this paper: categorical fragility is a measurable structural property that precedes observable data quality failure.

\section{Methodological Advantages}

The ISEC framework introduces several methodological contributions that distinguish it from traditional data quality assessment and string similarity approaches.

\subsection{Ex-Ante Structural Risk Modeling}

Conventional data quality methodologies operate reactively, identifying errors after they have been instantiated in the dataset \cite{Cichy2019,Wand1996,Wang1996}. In contrast, ISEC models \emph{latent structural vulnerability} prior to error occurrence \cite{Simard2019,unterkalmsteiner_2023}.

By evaluating inter-category separability under stochastic perturbation, 
ISEC transforms categorical fragility into a measurable ex-ante risk 
indicator \cite{Wang1996,Cichy2019}.

\subsection{Ordinal Sensitivity Rather Than Absolute Error Estimation}

ISEC does not attempt to estimate the true probability of error occurrence. 
Instead, it constructs an ordinal vulnerability ranking over category pairs \cite{Wang1996}.

This methodological decision provides two advantages:

\begin{itemize}
\item It avoids unrealistic probabilistic assumptions about user behavior \cite{Feit2016}.
\item It remains robust across heterogeneous domains and input contexts.
\end{itemize}

The index therefore functions as a decision-support prioritization 
instrument rather than a predictive classifier \cite{MendesSampaio2015}.

\subsection{Integration of Morphological, Semantic, and Frequency Dimensions}

Traditional edit-distance metrics focus exclusively on syntactic 
transformations \cite{Berger2021,Po2020,Navarro2001}. Semantic similarity measures, in isolation, fail to 
capture typographical perturbation dynamics \cite{Aliero2023}.

ISEC integrates three orthogonal dimensions:

\begin{itemize}
\item Morphological cost measure (via weighted edit operations \cite{Berger2021,Zhao2019}),
\item Semantic distance (via normalized similarity measures \cite{Johnson2024,Reimers2019}),
\item Empirical exposure (via frequency normalization).
\end{itemize}

This multi-dimensional formulation enables detection of structural 
ambiguity zones that would remain invisible under any single metric \cite{Velden2024}.

\subsection{Configurable Risk Encoding}

The use of character-level cost matrices and the penalization parameter 
$k$ allows organizations to encode empirical error tendencies and 
risk tolerance directly into the distance function \cite{Wang1996,Cichy2019}.

This transforms the metric from a static similarity computation into 
a governance-calibrated structural model \cite{Miller2025}.

Under default symmetric costs, ISEC reduces to a classical 
Damerau-Levenshtein framework \cite{Berger2021,Zhao2019}. Under customized matrices, it becomes 
a domain-aware categorical robustness evaluator.

\subsection{Domain Independence and Technological Sovereignty}

ISEC does not require proprietary software, pre-trained domain models, 
or external APIs. All components can be implemented using open-source 
libraries and local computational resources.

This property is particularly relevant for SMEs operating under budget or technology infrastructure constraints \cite{Ghasemaghaei2018,Gnther2019,Alsousi2022}.

\subsection{Alignment with Decision Support Systems}

Unlike record-level anomaly detection methods, ISEC operates at the 
taxonomic layer, where categorical aggregation directly influences 
Key Performance Indicators and dashboard outputs.

By identifying structurally fragile category configurations, the 
method contributes to Decision Support System robustness \cite{Shrestha2021} before 
distortions propagate into managerial metrics \cite{Cichy2019}.

In this sense, ISEC extends data quality assessment from operational 
correction toward structural decision-theoretic design \cite{Wang1996,Wand1996}.

\section{Discussion}

This work reframes categorical data quality as a problem of 
\emph{structural robustness under perturbation}. Rather than treating errors as isolated record-level anomalies, we conceptualize fragility as a geometric property of the taxonomy itself \cite{unterkalmsteiner_2023,Velden2024}.

\subsection{From Error Detection to Structural Design}

Most data quality frameworks focus on detection, correction, or post-hoc cleansing \cite{Cichy2019}. Such approaches implicitly assume that errors are identifiable once instantiated. However, the cases examined in 
this study demonstrate the existence of irrecoverable distortions that become statistically indistinguishable from valid entries \cite{Cichy2019}.

ISEC shifts attention from error occurrence to structural preconditions of ambiguity. By modeling inter-category proximity before perturbation occurs, the metric operates at the design layer of the classification system \cite{Wang1996,Cichy2019}.

This represents a transition from operational data hygiene to \emph{taxonomic engineering}.

\subsection{Decision-Theoretic Implications}

Decision Support Systems depend on aggregation over nominal categories \cite{Shrestha2021}. If categorical separability collapses, downstream Key Performance Indicators remain internally coherent but semantically distorted.

This creates a specific epistemic risk: 
\emph{structured misinformation}.

Unlike random noise, structured misinformation is stable under aggregation \cite{MendesSampaio2015,Shrestha2021}. It produces consistent dashboards, coherent reports, and plausible trends, while misrepresenting the underlying operational reality \cite{Ghasemaghaei2018,Velden2024}.

ISEC contributes to DSS robustness by identifying fragile taxonomic configurations before they propagate into 
managerial decisions.

\subsection{ISEC as a Governance Instrument}

The configurability of character-level cost matrices and the penalization parameter $k$ implies that categorical 
robustness is not purely technical \cite{Wang1996,Cichy2019}. Organizations implicitly define acceptable ambiguity thresholds whenever they choose naming conventions, abbreviations, or UI input constraints. ISEC makes these thresholds explicit and measurable \cite{Miller2025}.

\subsection{Boundary Conditions}

ISEC is designed for nominal categorical systems characterized by manual or semi-manual input processes. It is less applicable to: (i) Fully automated sensor-based categorical generation, (ii) Ontologically constrained taxonomies with enforced unique identifiers \cite{unterkalmsteiner_2023}, (iii) High-dimensional continuous feature spaces.

Additionally, ISEC provides ordinal vulnerability rankings 
rather than probabilistic error predictions \cite{Wang1996}. It does not model 
user cognition directly, nor does it infer behavioral error rates \cite{Feit2016}. 
Its contribution lies in structural susceptibility analysis and risk mitigation tool \cite{Wang1996,Cichy2019}.

\subsection{Beyond SMEs}

Although motivated by SME constraints \cite{Alsousi2022,GohZhiJi2023,Gnther2019}, the framework generalizes to any environment where; (i) Nominal categories drive aggregation,(ii) Manual input introduces perturbation such as speech to text (STT) or optical character recognition (OCR)\cite{Aliero2023},(iii) Taxonomies that evolve organically.

Public administration, healthcare coding systems, regulatory 
classification schemes, and inventory management platforms 
may all exhibit similar structural fragility \cite{Cichy2019}.

Thus, the relevance of ISEC extends beyond SMEs to broader 
information governance contexts \cite{Miller2025}.

\subsection{From Reactive Data Quality to Information Robustness}

Classical data quality frameworks conceptualize quality as conformance to specification after data instantiation \cite{Cichy2019,Wang1996}. Dimensions such as accuracy, completeness, and consistency evaluate records relative to predefined constraints \cite{Miller2025,Wand1996}.

ISEC suggests a complementary perspective: quality as structural robustness prior to error occurrence \cite{Simard2019,unterkalmsteiner_2023}. We define \emph{information robustness} as the degree to which a 
categorical system preserves semantic separability under bounded stochastic perturbation \cite{lorena_2019,unterkalmsteiner_2023}. Under this perspective; (i) Data accuracy evaluates correctness ex post \cite{Wang1996}, (ii) Information robustness evaluates separability ex ante \cite{unterkalmsteiner_2023}.

A taxonomy may contain no current errors and still exhibit low 
robustness \cite{unterkalmsteiner_2023,lorena_2019} if small perturbations are sufficient to collapse 
inter-category distance \cite{Berger2021,Po2020,Navarro2001}. Thus, robustness constitutes a preventive data quality dimension operating at the architectural level rather than the record level \cite{Wang1996,Cichy2019,Simard2019}.

ISEC operationalizes this dimension by quantifying the susceptibility 
of categorical geometry to distance compression under edit-based noise.

\section{Conclusion}

Categorical errors in SMEs are not merely operational nuisances but potential generators of structured misinformation within Decision Support Systems \cite{MendesSampaio2015,Shrestha2021}. Some of these errors are irrecoverable once embedded. The ISEC provides an ordinal, preventive, and scalable framework to measure categorical sensitivity prior to error manifestation \cite{lorena_2019,unterkalmsteiner_2023}. By modeling inter-category proximity and penalized transformation costs \cite{Navarro2001,Zhao2019}, it operationalizes a previously unmeasured dimension of information quality: \textbf{structural robustness}.

Several concrete research directions emerge from the current implementation. The most immediate limitation of ISEC is its reliance on manually defined cost matrices, which require expert knowledge of the specific input process (keyboard layout \cite{Feit2016}, OCR behavior, speech-to-text patterns). A natural extension is to automate matrix learning through a Hidden Markov Model (HMM) framework, where the observed characters typed by users constitute the emission sequence and the intended canonical categories constitute the hidden states. Applying the Viterbi algorithm over a corpus of confirmed corrections would allow the system to estimate character-level confusion probabilities empirically, updating cost assignments in a semi-supervised manner. This would enable the index to adapt continuously to evolving error tendencies without manual recalibration, a property particularly valuable for SMEs that cannot dedicate technical personnel to parameter tuning \cite{Gnther2019}.

A second extension concerns the dimensional scope of the index. The current formulation integrates morphological and semantic distances, but not the syntactic structure. For domains governed by formal naming conventions, such as ISO part codes, pharmaceutical identifiers, or financial instrument codes. A third syntactic dimension based on positional character importance or parse-tree similarity could improve discriminability in cases where two codes differ only in structurally significant positions. 

Separately, integrating ISEC as a component within the Gower distance framework for heterogeneous datasets \cite{Velden2024} would extend its applicability to mixed-type records, replacing the binary categorical treatment in Gower with a continuous sensitivity score. 

Finally, embedding ISEC as a native plugin within open-source data governance platforms such as OpenMetadata, DataHub or Amundsen would transform it from an analytical instrument into a continuous governance mechanism \cite{Miller2025}, enabling automated scoring, threshold-based alerting, and structural vulnerability tracking as part of an organization's standard data operations \cite{Wang1996,Cichy2019}.


\bibliographystyle{unsrt}


\bibliography{lib}

\end{document}